\documentclass[runningheads]{llncs}
\usepackage[T1]{fontenc}
\usepackage{graphicx, enumitem}
\usepackage{amsmath, bm}
\usepackage{pgfplots}
\pgfplotsset{compat=1.18}
 \usepackage{booktabs}
 \usepackage{hyperref}
\begin{document}
\title{Reliable Fusion of Conflicting Experts}
\author{Pranuthi Tenali\inst{1} \and 
Sahil Sidheekh\inst{1} \and
Saurabh Mathur\inst{3} \and
Vijayalakshmi Saravanan\inst{2} \and
Erik Blasch\inst{4} \and
Kristian Kersting\inst{3} \and
Sriraam Natarajan\inst{1}}
\authorrunning{Tenali et al.}
\institute{The University of Texas at Dallas \\
\email{\{pranuthi.tenali, sahil.sidheekh, sriraam.natarajan\}@utdallas.edu}\and
The University of Texas at Tyler \\
\email {vsaravanan@uttyler.edu}\and
TU Darmstadt \\
\email{saurabh.mathur@tu-darmstadt.de, kersting@cs.tu-darmstadt.de} \and
Air Force Research Lab \\
\email{erik.blasch.1@us.af.mil}}
\maketitle              % typeset the header of the contribution
\begin{abstract}

We study the problem of aggregating opinions from multiple black-box experts in noisy, conflict-prone settings where expert reliability varies across inputs. Static aggregation methods, such as majority voting, fail to capture this variability and often yield unreliable outcomes under disagreement. We propose a tractable, probabilistic-circuit-based fusion framework that dynamically combines expert responses using context-specific credibility estimates, enabling principled and reliable reasoning. The framework is agnostic to the underlying experts and does not require access to their internal representations or any retraining. We empirically validate our approach on multiple-choice question answering tasks using multiple LLMs as experts, comparing against individual models and static ensemble baselines. Our method consistently improves predictive performance and produces more reliable decisions under conflict, highlighting the effectiveness of context-aware credibility modeling for robust multi-expert fusion.

\end{abstract}

\section{Introduction}

Aggregating information from multiple sources is a foundational challenge in Artificial Intelligence and Dynamic Data-Driven Application Systems (DDDAS)~\cite{dddas,Darema2023}. Agents operating in complex environments, such as medicine, rarely pursue a single monolithic objective; instead, they must balance multiple conflicting criteria whose relative importance and relevance shift dynamically based on the current situation~\cite{natarajan2005dynamic}. For instance, a clinical decision-support agent must dynamically shift its objective from identifying rare pathologies to optimizing patient stabilization when vitals begin to deteriorate rapidly. To navigate this variability, the agent must rely on diverse streams of information, including distributed sensor arrays, specialized knowledge bases, and, more generally, a panel of experts. 
In an ideal setting, combining judgments from a panel of experts would yield results better than any single source. However, real-world experts rarely possess uniform knowledge; instead, their reliability varies drastically across different regions of the problem space~\cite{abels2023expertise}.

When these specialized experts inevitably disagree, simple fusion methods such as statically weighted averaging prove inadequate, because they make the highly restrictive assumption that each expert possesses a fixed, global reliability across the entire problem space. This rigid formulation treats reliability as an intrinsic property of the expert rather than a dynamic interaction between the expert's specialization and the problem context. Formally resolving such disagreements requires a mechanism capable of computing instance-level trade-offs, separating a source's static structural importance from its dynamic, context-specific reliability~\cite{smarandache2010fusion}.

This aggregation problem has become particularly relevant with the rise of large language models (LLMs) deployed as multi-agent systems and specialized prompt ensembles~\cite{xu2023expertprompting,do2024multi}. Modern AI pipelines frequently orchestrate networks of proprietary black-box models acting under targeted personas, such as specialized agents for medicine, physics, and computer science. Since these systems are accessed exclusively through commercial APIs, their internal representations are completely hidden from the user. This poses an architectural challenge for traditional dynamic gating solutions, such as simple mixture-of-experts (MoE) models that struggle to reason about context-specific conflicts without fine-tuning each expert's internal representations.

Dealing with heterogeneous inputs having different and contextual reliability is also central to DDDAS systems that must fuse information from multiple sources while adapting to dynamic environments.
% depending on the context.
\begin{figure}[!t]
    \centering
    \includegraphics[width=\linewidth]{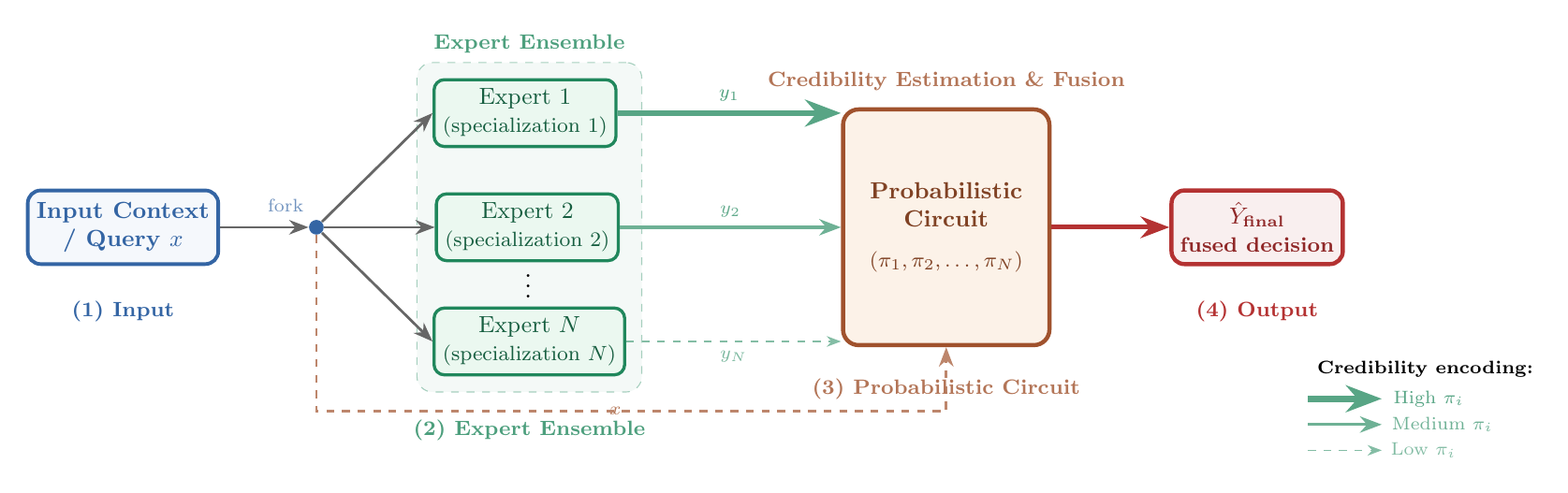}
    \caption{Our framework dynamically aggregates opinions from an ensemble of black-box experts with distinct, varying specializations. Given an input query $x$, the context is broadcast to an ensemble of $N$ black-box experts with distinct specializations, each producing a response $y_i$. A probabilistic circuit receives all expert responses alongside the input context and computes context-specific credibility weights $(\pi_1,\dots,\pi_N).$ The weighted combination yields a fused decision $\hat{Y}_\text{final}.$}
    \label{fig:placeholder}
\end{figure}
% \textbf{[What is our solution?]}
We address this challenge by introducing a lightweight expert fusion framework that dynamically aggregates conflicting black-box expert opinions using probabilistic circuits. By treating the input query text as a context variable, our framework learns to estimate the instance-level reliability and situational weight of each expert. This enables exact computation of context-specific credibility scores that provide a measure of how much each source can be trusted, thus improving the explanability within the DDDAS feedback loop. We make the following key contributions: (1) We propose a lightweight framework using probabilistic circuits to dynamically aggregate conflicting black-box expert opinions based on input-specific credibility estimates.
%\textbf{[How do we know that it works?]}
(2) We demonstrate on the MMLU question answering benchmark that our framework consistently outperforms static ensembles, accurately resolving expert disagreement under intense conflict.

\section{Background} 
Our framework reasons about multiple conflicting expert opinions by combining multi-source fusion and multi-criteria optimization with conditional probabilistic circuits. We review these foundational concepts in this section.

\subsection{Expert fusion}

The problem of aggregating predictions from a panel of $N$ heterogeneous experts is formalized by treating each expert as a distinct information source that provides a localized opinion or prediction vector $y_i$ for a given query $x$. The fusion system must reason about these individual predictions to construct a single, unified consensus decision $Y_\text{agg}.$ A simple way for achieving this is the statically weighted average \cite{natarajan2005learning}, expressed as:
$Y_\text{agg} = \sum_{i=1}^N w_i y_i,$
where $w_i\in[0,1]$ is a fixed reliability coefficient assigned to expert i, such that $\sum_i w_i = 1.$ These weights are learned by optimizing the ensemble performance over a small validation dataset, and kept fixed at inference.

However, in complex environments, an agent's optimal behavior cannot be dictated by a single, monolithic goal; instead, the relative importance of conflicting criteria shifts dynamically based on situational states. Multi-criteria learning systems reason about distinct specialized behaviors and dynamically interpolate between them using situational weight vectors. Concretely, this transforms the fixed weight vector into a dynamic, conditional function $w_i(x)$ of the problem context $x$. Another approach is to explicitly partition the global problem space into discrete, localized sub-regions using \textit{expertise trees} \cite{abels2023expertise}. This model assigns the experts high reliabilty within their specialized sub-region, but significantly reduced it when the context $x$ crosses into a neighboring partition managed by a different specialization. Both approaches are unified by the context-specific credibility-aware multimodal fusion (C$^2$MF) framework \cite{tenali2026context}, which models source combination through a conditional probabilistic circuit -- an architecture that subsumes weighted means and expertise trees as special cases. We adopt C$^2$MF as the foundation of our approach and extend it to the black-box multi-LLM setting in Section~\ref{sec:method}.

\subsection{Fusion via Probabilistic Circuits}
A Probabilistic Circuit (PC)~\cite{Choi20,sidheekh2024building} represents a joint distribution $P(\mathbf{X})$ as a computational graph. Specifically, it is a rooted DAG whose leaves are univariate distributions and whose internal nodes are sum nodes (weighted mixtures) or product nodes (factorizations over disjoint variable subsets). Their chief advantage over general neural models is that PCs support the exact computation of marginal and conditional probabilities in time linear in circuit size~\cite{SPNPoon2011}, under two structural constraints. A PC is \textit{smooth} if every sum node's children share the same scope, and \textit{decomposable} if every product node's children have disjoint scopes. Together, smoothness and decomposability ensure that marginal and conditional queries can be answered in a single bottom-up 
pass over the circuit. Since smooth and decomposable PCs are differentiable computational graphs, their parameters can be learned via backpropagation. While their structure can also be learned from data, in practice, random tensorized structures~\cite{peharz20a-rat-spn} are used.

The credibility-aware multimodal fusion framework (CMF~\cite{sidheekh2025credibility,sidheekh2024robustness}) uses PCs to combine predictions from multiple unimodal predictors. In it, the PC models the joint distribution over predictions from each source modality $\{p_i\}_{i=1}^{N}$ and target $Y$. The fused decision is obtained by exact conditional inference: $P(Y \mid p_1, \dots, p_N)$. 

Beyond direct fusion, the PC's tractability allow the \textit{credibility} of each source to be quantified explicitly. The credibility of source $i$ is measured by computing the KL-divergence between the full posterior $P(Y \mid p_1, \dots, p_N)$ and the posterior obtained by marginalizing $p_i$ out of the circuit. These credibility scores can in turn be used to form a credibility-weighted mean (CWM) prediction.

\subsection{Context-specific fusion via Conditional Probabilistic Circuits} 
A key limitation of CMF is that the the credibility assigned to each source  depends only on its predicted values $\{p_i\}$ and not on the input  context. In complex environments where expert reliability varies across  the problem space, this can lead to unreliable decisions  under disagreement.  The context-specific credibility-aware multimodal fusion (C$^2$MF)  framework~\cite{tenali2026context} addresses this by replacing the  static PC with a Conditional Probabilistic Circuit (CPC), whose  sum-node weights are predicted by a neural gate conditioned on an  external context variable $z$~\cite{shao2020conditional}. This yields  a tractable model of $P(Y, \mathbf{p} \mid  z)$ whose fusion logic  adapts to each input. The credibility of source $i$ under context $z$  is then computed by the same marginalization procedure as in CMF, but  now conditioned on $z$: $$ c_i(\mathbf{p}, z) = D_{KL}\!\left(P(Y \mid \mathbf{p}, z)\ \|\  P(Y \mid \mathbf{p} \setminus p_i, z)\right), $$ yielding context-specific information credibility (CSIC) scores $\overline{c}_i(\mathbf{p}, z)$  that reflect each source's situational reliability given both its  prediction and the current input context.

\section{Fusing conflicting Expert Opinions}\label{sec:method}

Agents operating in complex dynamic environments must select from  conflicting experts based on their current situation. We study this  setting where $K$ black-box LLM experts with distinct specializations are queried on multiple-choice questions, and their responses must be aggregated into a single reliable prediction without 
access to their internal representations. We formalize this as:
\fbox{
\parbox{.95\columnwidth}{
\textbf{Given:} A small labeled dataset of $N$ query-answer pairs, and responses $\{p_k\}_{k=1}^{K}$ (probability vectors over $C$ answer choices) from $K$ black-box experts that may disagree, with reliability varying across queries.\\[4pt]
\textbf{To Do}: Learn a tractable aggregation model $\mathcal{M}$ that produces reliable predictions on unseen queries
}
}

We adapt the C$^2$MF framework~\cite{tenali2026context} to the setting of black-box LLM experts on multiple-choice question answering. The framework operates in two stages: encoding the query context and fusing the expert responses via a Conditional Probabilistic Circuit.
Each expert $k \in \{1, \dots, K\}$ receives a query $x$ and returns a probability vector $p_k$ over $C$ answer choices.

The query $x$ is encoded into a fixed-dimensional context vector  $z \in \bm{R}^d$ using a pretrained sentence embedding model.  This embedding captures the semantic content of the query and serves  as the conditioning variable for the CPC, allowing the fusion function to adapt to the nature of each question without requiring  access to expert internals.

The expert responses $\mathbf{p} = (p_1, \dots, p_K)$ and the  context vector $z$ are passed to a CPC that models the joint  distribution $P(Y \mid \mathbf{p}, z)$ over the correct answer $Y$.  The sum-node weights of the CPC are predicted by a small MLP  conditioned on $z$, so that the circuit's fusion logic adapts to  each query. The conditional PC is trained by maximizing the conditional log-likelihood over a small labeled dataset of $N$ query-answer pairs.

At inference time, the context-specific credibility of each expert  is computed via the CPC's tractable marginalization routines: 
$$ c_k(\mathbf{p}, z) = D_{KL}\!\left(P(Y \mid \mathbf{p}, z)\ \|\  P(Y \mid \mathbf{p} \setminus p_k, z)\right). $$ 
These scores are normalized to obtain a credibility distribution  $\overline{c}_k(\mathbf{p}, z) \in [0,1]$ with $\sum_k  \overline{c}_k = 1$. The final prediction is a credibility-weighted  mean over expert responses: 
$ \hat{Y} = \sum_{k=1}^{K} \overline{c}_k(\mathbf{p}, z)\, p_k.$
Experts whose predictions align poorly with the fused posterior  under the current context receive low credibility and are  down-weighted automatically.

 \begin{table}[t]
\centering
\setlength{\tabcolsep}{3pt}
\caption{Mean Test Performance on MMLU across individual subject experts (Medical, CS, Physics), the Single-Oracle Expert (SOE) skyline, the Random Expert Selection (RE) baseline, and fusion methods, averaged over $3$ trials. Fusion methods include non-context-aware approaches (WM, Noisy-OR, DPC, CWM), and context-aware approaches (C$^2$DPC and C$^2$WM). The context-aware methods are competitive across all the metrics, with C$^2$WM achieving the highest Precision and Recall, and C$^2$DPC the highest AUROC while additionally yielding interpretable per-expert context-specific credibility scores.}
\label{tab:experiment_metrics}
\begin{tabular}{lccccc}
\toprule
{\textbf{Experiment}} &
{\textbf{Accuracy}} &
{\textbf{AUROC}} &
{\textbf{Precision}} &
{\textbf{Recall}} &
{\textbf{F1 Score}} \\
\midrule
{\small Medical}          & {$55.40 \scriptstyle\pm 7.20$} & {$76.11 \scriptstyle \pm 3.69$} & {$57.22 \scriptstyle \pm 7.02$} & {$55.40 \scriptstyle \pm 7.20$} & {$54.87 \scriptstyle \pm 8.42$} \\
{\small CS} & {$ 62.13 \scriptstyle \pm 5.62$} & {$78.35 \scriptstyle \pm 3.40$} & {$62.47 \scriptstyle \pm 6.28$} & {$62.13 \scriptstyle \pm 5.62$} & {$61.87 \scriptstyle \pm 5.79$} \\
{\small Physics}          & {$ 60.29 \scriptstyle\pm 7.38$} & {$ 75.74 \scriptstyle\pm 5.68$ }& {$57.22 \scriptstyle\pm 7.02$} & {$ 60.29 \scriptstyle\pm 7.38$} & {$ 59.83 \scriptstyle\pm 7.55$} \\
\midrule
{SOE}       & {$ 67.16 \scriptstyle\pm 2.04$} & {$ 80.19 \scriptstyle\pm 0.94$} & {$ 68.05 \scriptstyle\pm 2.15$} & {$ 67.16 \scriptstyle\pm 2.04$} & {$ 66.59 \scriptstyle\pm 2.45$} \\
{RE}       & {$ 62.62 \scriptstyle\pm 2.89$} & {$ 79.16 \scriptstyle\pm 1.72$} & {$ 63.18 \scriptstyle\pm 3.11$} & {$ 62.62 \scriptstyle\pm 2.89$} & {$ 62.48 \scriptstyle\pm 3.18$} \\
{NoisyOR}      & {$ 65.93 \scriptstyle\pm 2.99$} & {$ 84.30 \scriptstyle\pm 1.82$} & {$ 67.63 \scriptstyle\pm 2.45$} & {$ 67.06 \scriptstyle\pm 2.87$} & {$ 65.35 \scriptstyle\pm 3.16$} \\
{WM} & {$ 66.54 \scriptstyle\pm 4.09$} & {$ 84.31 \scriptstyle\pm 1.50$} & {$ 68.23 \scriptstyle\pm 3.65$} & {$ 67.66 \scriptstyle\pm 4.26$} & {$ \mathbf{66.03 \scriptstyle\pm 4.20}$} \\
{CWM}          & {$ 65.93 \scriptstyle\pm 2.50$} & {$ 84.55 \scriptstyle\pm 2.00$} & {$ 67.97 \scriptstyle\pm 2.08$} & {$ 67.05 \scriptstyle\pm 2.56$} & {$ 65.27 \scriptstyle\pm 2.52$} \\
{DPC}          & {$ 64.46 \scriptstyle\pm 3.34$} & {$ 82.48 \scriptstyle\pm 1.56$} & {$ 67.49 \scriptstyle\pm 4.04$} & {$ 65.64 \scriptstyle\pm 3.12$} & {$ 63.82 \scriptstyle\pm 3.57$} \\
{C$^2$DPC}         & {$ 64.83 \scriptstyle\pm 2.66$} & {$ \mathbf{85.06 \scriptstyle\pm 1.27}$} & {$ 66.09 \scriptstyle\pm 2.76$} & {$ 65.84 \scriptstyle\pm 2.35$} & {$ 64.35 \scriptstyle\pm 2.87$} \\
{C$^2$WM}         & {$ \mathbf{66.54 \scriptstyle\pm 3.51}$} & {$ 84.28 \scriptstyle\pm 1.45$} & {$ \mathbf{68.43 \scriptstyle\pm 3.39}$} & {$ \mathbf{67.75 \scriptstyle\pm 3.58}$} & {$ 65.93 \scriptstyle\pm 3.45$ }\\
\bottomrule
\end{tabular}
\end{table}
\section{Empirical Evaluation}
We aim to answer the following two key research questions.
\begin{itemize}
\item \textbf{(Q1)} Does expert fusion via C$^2$MF produce more accurate answers?
\item \textbf{(Q2)} Does C$^2$MF improve expert calibration?
\end{itemize}

\textbf{Dataset} MMLU \cite{hendrycks2021measuring} is a multitask language understanding benchmark designed to measure a language model's performance across diverse tasks and domains. The dataset consists of multiple-choice question-answer-subject entries spanning 57 different tasks. In this work, we choose a subset of $10$ tasks and cluster them into $3$ expert domains: Medicine, Physics, and Computer Science. 

\textbf{Prompt Generation} We generate prompts using the Expert Prompting mechanism \cite{do2024multi} to elicit the subject-specific expert behaviour from the model. Specifically, for each topic, we sample a small set of examples as candidate set to identify the releavant expert roles and capabilities. We then prompt another language model for the top $3$ expert roles best suited to answer each question effectively. Then, we aggregate the roles across all the examples in the candidate set and query the model to get a more general expert that encapsulates most, if not all, of these expert capabilities. 

\textbf{Querying the expert} Given the expert prompt and the question, we query the model for the probability of each answer choice being correct. The resulting probability scores are averaged across $2$ trials to obtain the final predictions for that expert. We used gemini-2.5-flash-lite for our evaluation.
% , which are then fused to get multi-expert predictions.

\textbf{Methods} We evaluate the two C$^2$MF-based fusion functions - C$^2$DPC and C$^2$WM.\emph{C$^2$DPC} uses the CPC to model the $P(Y, p_1,\dots p_K\mid z)$. The final predictive distribution is obtained as $ P(Y\mid p_1,\ldots,p_K, z) = \frac{P(Y,p_1\dots p_K \mid z)}{P(p_1,\ldots,p_K \mid z)}$ where $p_i$ is the predictions from $i^{th}$ expert. \emph{C$^2$WM} combines the expert predictions as a weighted average using the context-specific credibility scores computed from a CPC as weights.  In addition, we consider the 4 context-agnostic baselines.

\begin{enumerate}
    \item \emph{Direct-PC (DPC)} is a context-agnostic, PC based combination function that models the joint distribution over the expert predictions and the label as $Y$, $P(Y, \bm{p})$. The final predictive distribution is then obtained by marginalization as $ P(Y\mid \bm{p}) = \frac{P(Y,\bm{p})}{P(\bm{p})}$
    \item \emph{Credibility Weighted Mean (CWM)} uses credibility scores derived from the CMF framework as weights to compute a weighted average of expert outputs.
    \item \emph{Weighted Mean (WM)} fuses the expert predictions $\bm{p}$ using a convex combination as $ P(Y|\bm{p}, X_K) = \sum_{i=1}^{K} w_i P(Y|p_i) $ where $w_i$ are learnable weights such that $0 \leq w_i \leq 1$ and $\sum_{i=1}^{K}w_i=1$.
    \item \emph{Noisy-OR} combines the expert predictions as $P(Y|\bm{p}) = 1 - \prod_{i=1}^K (1-P(Y|p_i)).$

\end{enumerate}

In addition to the individual subject expert and the fusion functions, we consider two additional baselines. \emph{Single-Oracle Expert (SOE)} is a skyline in which the prediction of the respective experts is considered the final prediction, providing an upper bound on the performance attainable under ideal conditions. \emph{Random Expert Selection (RE)} is a baseline in which one expert is sampled uniformly at random, and its prediction is used as the final prediction. The experimental details and code are available \href{https://github.com/Pranuthi23/cs_cred_llms}{here}.
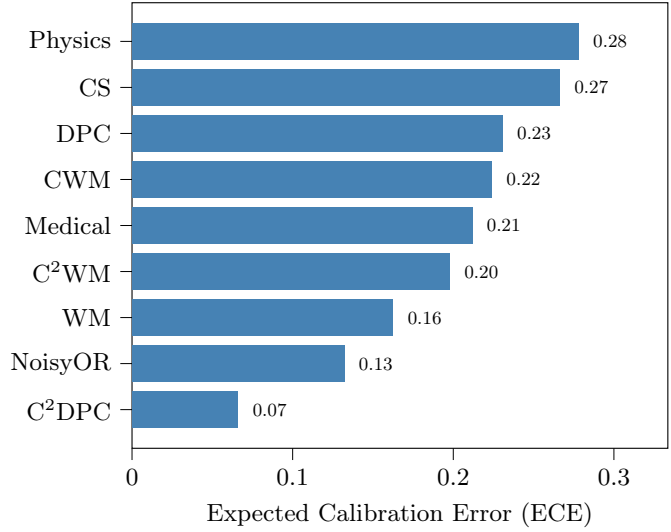
\begin{figure}[t]
    \centering
    \resizebox{.75\linewidth}{!}{
    % This file was created with tikzplotlib v0.10.1.
\begin{tikzpicture}

\definecolor{darkgray176}{RGB}{176,176,176}
\definecolor{steelblue}{RGB}{70,130,180}

\begin{axis}[
tick align=outside,
tick pos=left,
x grid style={darkgray176},
xlabel={Expected Calibration Error (ECE)},
xmin=0, xmax=0.333576732873917,
xtick={0,0.1,0.2,0.3},
xtick style={color=black},
y grid style={darkgray176},
ylabel={},
ymin=-0.84, ymax=8.84,
ytick style={color=black},
ytick={0,1,2,3,4,5,6,7,8},
yticklabels={C$^2$DPC,NoisyOR,WM,C$^2$WM,Medical,CWM,DPC,CS,Physics}
]
\draw[draw=none,fill=steelblue] (axis cs:0,-0.4) rectangle (axis cs:0.0659137902160486,0.4);
\draw[draw=none,fill=steelblue] (axis cs:0,0.6) rectangle (axis cs:0.132083013653755,1.4);
\draw[draw=none,fill=steelblue] (axis cs:0,1.6) rectangle (axis cs:0.162483776609103,2.4);
\draw[draw=none,fill=steelblue] (axis cs:0,2.6) rectangle (axis cs:0.197686195373535,3.4);
\draw[draw=none,fill=steelblue] (axis cs:0,3.6) rectangle (axis cs:0.211686208844185,4.4);
\draw[draw=none,fill=steelblue] (axis cs:0,4.6) rectangle (axis cs:0.2239737311999,5.4);
\draw[draw=none,fill=steelblue] (axis cs:0,5.6) rectangle (axis cs:0.230579788486163,6.4);
\draw[draw=none,fill=steelblue] (axis cs:0,6.6) rectangle (axis cs:0.266214713454247,7.4);
\draw[draw=none,fill=steelblue] (axis cs:0,7.6) rectangle (axis cs:0.277980610728264,8.4);
\draw (axis cs:0.0709137902160486,0) node[
  scale=0.75,
  anchor=west,
  text=black,
  rotate=0.0
]{0.07};
\draw (axis cs:0.137083013653755,1) node[
  scale=0.75,
  anchor=west,
  text=black,
  rotate=0.0
]{0.13};
\draw (axis cs:0.167483776609103,2) node[
  scale=0.75,
  anchor=west,
  text=black,
  rotate=0.0
]{0.16};
\draw (axis cs:0.202686195373535,3) node[
  scale=0.75,
  anchor=west,
  text=black,
  rotate=0.0
]{0.20};
\draw (axis cs:0.216686208844185,4) node[
  scale=0.75,
  anchor=west,
  text=black,
  rotate=0.0
]{0.21};
\draw (axis cs:0.2289737311999,5) node[
  scale=0.75,
  anchor=west,
  text=black,
  rotate=0.0
]{0.22};
\draw (axis cs:0.235579788486163,6) node[
  scale=0.75,
  anchor=west,
  text=black,
  rotate=0.0
]{0.23};
\draw (axis cs:0.271214713454247,7) node[
  scale=0.75,
  anchor=west,
  text=black,
  rotate=0.0
]{0.27};
\draw (axis cs:0.282980610728264,8) node[
  scale=0.75,
  anchor=west,
  text=black,
  rotate=0.0
]{0.28};
\end{axis}

\end{tikzpicture}
    }
    \caption{Expected Calibration Error (ECE) on the test set across individual subject experts (Physics, CS, Medical) and different fusion methods averaged over $3$ trials. Fusion methods include non-context-aware approaches (WM, Noisy-OR, DPC, CWM), and context-aware approaches (C$^2$DPC, C$^2$WM). The context-aware C$^2$DPC method is the best calibrated, achieving the lowest ECE.}
    \label{fig:ece}
\end{figure}

\textbf{Q1. Predictive Performance}  Table \ref{tab:experiment_metrics} presents the performance of the methods considered. Overall, fusing the predictions outperforms the individual subject experts, demonstrating the benefit of the wisdom of crowds compared to individual experts alone. Moreover, both the context-aware fusion methods are performing better than their respective static counter parts in terms of all the classification metrics. This shows that exploiting the contextual information for achieving better performance and increased reliability.

\textbf{Q2. Calibration} To more rigorously assess the calibration of the models, we compute the Expected Calibration Error (ECE), which measures the
average gap across the model’s predicted confidence and observed accuracy. It is calculated by dividing the range of predicted probabilities into $M$ distinct bins $B_1, \dots B_M$, and computed as 
\begin{equation*}
    \text{ECE} = \sum_{i=1}^{M} \frac{|B_i|}{N} \left| \text{acc}(B_i) - \text{conf}(B_i) \right|
\end{equation*}
where $|B_i|$ indicates the number of predictions in the $i$-th bin, $\text{acc}(B_i)$ is the empirical accuracy of predictions within that bin, and $\text{conf}(B_i)$ is the mean predicted confidence of the bin. Lower ECE values indicate better calibration. \ref{fig:ece} shows that C$^2$DPC achieves least ECE of all the fusion methods, indicating higher calibration. Furthermore, the context-aware fusion methods exhibit lower ECE than their context-agnostic counterparts. These results suggest that the C$^2$MF framework not only improves predictive performance but also produces more reliable and calibrated confidence estimates.

\section{Conclusion}

We considered the challenging problem of fusing information from conflicting experts. To this effect, we proposed a tractable, probabilistic approach to fusion that aggregates multiple experts in a principled and seamless manner. We empirically show that our framework achieves better performance and calibration over individual experts as well as other fusion functions. Since ours is a first, early attempt in this direction, there are several different ways of extending this work. First, one could use diffusion models for aggregating the different models, and consider fusing information from different types of experts, namely both human and AI experts. In addition, there is a critical need to move beyond multiple-choice questions. Exploring causal question answering and counterfactual reasoning are extremely interesting directions while training experts from the consensus remains an exciting direction for future research.

\section{Acknowledgements}
The authors gratefully acknowledge the support from AFOSR Award FA9550-23-1-0239, DOE FAIR Award DE-SC0026194, and the Cluster of Excellence “Reasonable AI” funded by the German Research Foundation (DFG) under Germany’s Excellence Strategy (EXC-3057).
\bibliographystyle{splncs04}
\bibliography{ref}

\end{document}